%% file: iclr2026.tex
\documentclass{article} 
\usepackage{iclr2026_conference,times}

\input{math_commands.tex}

\usepackage{hyperref}
\usepackage{url}
\usepackage{graphicx}
\usepackage{amsmath}
\usepackage{amssymb}
\usepackage{booktabs}
\usepackage{multirow}
\usepackage{natbib}
\usepackage{bibentry}
\usepackage{caption}
\usepackage{algorithm}
\usepackage{algorithmic}
\usepackage{xcolor}
\usepackage{wrapfig}
\usepackage{array}
\usepackage{enumitem}
\usepackage{colortbl}
\newcommand{\tabref}[1]{Table~\ref{#1}}
\newcommand{\equref}[1]{Equation~\ref{#1}}

\newcommand{\tablestyle}[2]{\setlength{\tabcolsep}{#1}\renewcommand{\arraystretch}{#2}\centering}
\newcommand{\green}[1]{\textcolor{green!60!black}{#1}}

\newcolumntype{g}{>{\columncolor{gray!30}}c}

\newcommand{\myparagraph}[1]{\noindent\textbf{#1.}}

\title{Enhancing Object Discovery for Unsupervised Instance Segmentation and Object Detection}

\author{Xingyu Feng\textsuperscript{\rm 1}\thanks{Equal contribution.} \quad Hebei Gao\textsuperscript{\rm 2}\footnotemark[1] \quad \textbf{Hong Li}\textsuperscript{\rm 1}\thanks{Corresponding author.} \\
\textsuperscript{1} College of Computer Science and Artificial Intelligence, Wenzhou University \\
\textsuperscript{2} Oujiang Laboratory (Zhejiang Lab for Regenerative Medicine, Vision and Brain Health), \\ \textsuperscript{$\;\;$} Eye Hospital, Wenzhou Medical University \\
\texttt{24451352011@stu.wzu.edu.cn \quad lihong@wzu.edu.cn}
}

\iclrfinalcopy 
\newif\ifappendixonly

\begin{document}
\ifappendixonly
\nobibliography{iclr2026}
\input{appendix}
\else

\maketitle

\begin{abstract}
We propose \textbf{C}ut-\textbf{O}nce-and-\textbf{LE}a\textbf{R}n (COLER), a simple approach for unsupervised instance segmentation and object detection. COLER first uses our developed CutOnce to generate coarse pseudo labels, then enables the detector to learn from these masks. CutOnce applies Normalized Cut (NCut) only once and does not rely on any clustering methods (e.g., K-Means), but it can generate multiple object masks in an image.
\textit{Our work opens a new direction for NCut algorithm in multi-object segmentation.}
We have designed several novel yet simple modules that not only allow CutOnce to fully leverage the object discovery capabilities of self-supervised model, but also free it from reliance on mask post-processing.
During training, COLER achieves strong performance without requiring specially designed loss functions for pseudo labels, and its performance is further improved through self-training. COLER is a zero-shot unsupervised model that outperforms previous state-of-the-art methods on multiple benchmarks. We believe our method can help advance the field of unsupervised object localization.
Code is available at: \url{https://github.com/Quantumcraft616/COLER}.
\end{abstract}

\section{Introduction}
In computer vision, segmentation tasks heavily rely on large-scale manual annotations, which significantly limits the development speed of the field. To reduce dependence on labeled data, researchers have begun exploring weakly-supervised or unsupervised approaches~\citep{unsupervised}.
This work focuses on exploring how to perform unsupervised instance segmentation and object detection efficiently. Found~\citep{Found} established the two-stage framework for unsupervised segmentation: generating pseudo labels followed by training a detector using them.

In recent years, methods that use self-supervised models to extract image features and then apply Normalized Cut (NCut)~\citep{NCut} to generate pseudo labels have made impressive progress, outperforming many other approaches. TokenCut~\citep{TokenCut}, MaskCut~\citep{CutLER}, and VoteCut~\citep{CuVLER} all use the DINO~\citep{DINO} model to extract features and produce reliable pseudo labels, achieving significant advances in unsupervised object localization field. DiffCut~\citep{DiffCut} employs a diffusion UNet~\citep{UNet} encoder for feature extraction, targeting unsupervised semantic segmentation.

The NCut paper~\citep{NCut} recommends two approaches for extending from single-object to multi-object segmentation. 
One approach is to perform NCut once, using the top-$n$ eigenvectors as an $n$-dimensional indicator vector, followed by clustering (e.g., K-Means) to segment multiple objects. VoteCut~\citep{CuVLER} adopts a clustering-based approach and generates masks using only the second smallest eigenvector. However, clustering methods require specifying the number of clusters, which reduces the generality of such approaches.
The other approach is to recursively partition the separated groups, that is, to further split the current foreground or background regions. MaskCut~\citep{CutLER} adopts this strategy by recursively partitioning the background. However, this approach clearly suffers from error accumulation as the number of recursive steps increases. 
These two types of methods have low computational efficiency, hindering their application and further exploration in related fields.

This paper uses NCut to segment multiple objects, but it differs from the two types of methods mentioned above. \textit{Our CutOnce neither relies on multiple applications of NCut nor on clustering methods when discovering multiple objects.} 
In short, by applying NCut only once, CutOnce can 
\begin{wraptable}{l}{0.53\textwidth}
	\centering
	\small
	\tablestyle{0.5pt}{1.0}
	\caption{\textbf{Key properties of our CutOnce and COLER with state-of-the-art methods.}}
	\begin{tabular}{lcc|c} 
			\toprule
			Train-Free Mask Generators & MaskCut & VoteCut & CutOnce \\ 
			\midrule
			Normalized Cut \#nums & 3  & 1  & 1   \\
			clustering method & \texttimes  &   \checkmark     & \texttimes   \\
			post-process
			mask  & \checkmark  & \checkmark  & \texttimes   \\
			self-supervised \#models & 1  & 6  & 1   \\
			max \#objects detected & 3  & 10 & \textbf{$>$10} \\
			mask generation time (s/img)  &  5.6   &   2.4    &  \textbf{0.23}    \\
			\midrule \midrule
			Pseudo Mask Learners & CutLER & CuVLER & COLER  \\ 
			\midrule
			pseudo mask loss function & \checkmark & \checkmark & \texttimes \\
			AP$_{50}^\text{mask}$ on COCO val2017 & 18.9 & 19.3 & \textbf{22.1} \\
			\bottomrule
	\end{tabular}
	\label{tab:methods_properties}
\end{wraptable}
discover multiple objects from an image rather than just one, which is the origin of its name.
Furthermore, \textit{CutOnce is capable of detecting masks for over 10 objects (examples are shown in the Appendix), which exceeds the maximum number of detectable targets of currently known methods.}
CutOnce does not require computationally expensive methods such as Conditional Random Field (CRF)~\citep{CRF}.
\tabref{tab:methods_properties} summarizes the key properties of CutOnce and popular existing methods~\citep{CutLER,CuVLER}, showing that CutOnce not only detects more objects but also generates annotations up to 10$\times$ faster.
COLER uses CutOnce's annotations for training and achieves good performance through self-training, without the need to design specialized loss functions to handle errors in the ``ground truth" provided by pseudo masks.

The contributions of this paper can be summarized as follows:
\textit{1)} We develop an efficient tool, CutOnce, for generating coarse masks. \textit{We introduce a novel paradigm for applying NCut to multi-object segmentation.}
\textit{2)} We train a detector, COLER, using pseudo masks generated by CutOnce. COLER is a zero-shot model that outperforms prior work across multiple datasets.

\section{Related Work}
\myparagraph{Self-Supervised Vision Transformer}
Self-supervised models are capable of learning deep features without human annotations or supervision. ViT~\citep{ViT} captures long-range dependencies between different regions in images through a global self-attention mechanism, making it easier to focus on semantically consistent target regions compared to CNN. DINO~\citep{DINO} combines both advantages, playing a key role in advancing unsupervised object localization. It adopts a teacher-student training framework and introduces a novel contrastive learning strategy that compares features from the original image and its random crops to learn stronger visual representations.
Thanks to the built-in spatial attention mechanism of the ViT architecture, DINO's attention maps can be directly used for localization and have shown advantages over previous methods. By further processing DINO's attention maps, more precise object regions can be obtained. And many methods~\citep{TokenCut, CutLER, CuVLER, CutS3D} have been developed based on this idea, achieving significant progress in their respective fields.

\myparagraph{Unsupervised Instance Segmentation and Object Detection}
LOST~\citep{LOST} is the first to localize objects by leveraging the final layer CLS token from a pre-trained transformer DINO~\citep{DINO} and computing patch-wise similarity within single image, but it can only localize one object.
MOST~\citep{MOST} extends this to multiple objects through entropy-based box analysis and clustering.
FreeSOLO~\citep{FreeSolo} uses features from DenseCL~\citep{DenseCL} to generate a set of ``queries" and ``keys" which are convolved to produce masks and also supports multiple object localization.
These methods do not rely on NCut and their robustness on large-scale datasets remains unconvincing.

TokenCut~\citep{TokenCut} is the first to apply NCut to features extracted by DINO, significantly improving the quality of pseudo labels, but it can only segment single instance. CutLER~\citep{CutLER} recursively applies NCut on one image to generate masks for multiple instances. CuVLER~\citep{CuVLER} uses multiple self-supervised models to generate diverse mask proposals and selects the best masks through clustering and pixel-wise voting. In addition, it assigns a confidence score to each pseudo mask.
\tabref{tab:methods_properties} presents the properties of the above methods and compares them with ours. CutS3D~\citep{CutS3D} introduces 3D information to enhance segmentation in 2D images, showing certain advantages in handling overlapping or connected objects and demonstrating strong potential for real-world applications. DiffNCut~\citep{DiffNCut} proposes Differentiable Normalized Cuts. In other words, it uses NCut to propagate gradients and fine-tune DINO.
Since NCut enhances object discovery, methods that use it outperform those that do not.

\begin{figure}[t]
	\centering
	\includegraphics[width=1.0\linewidth]{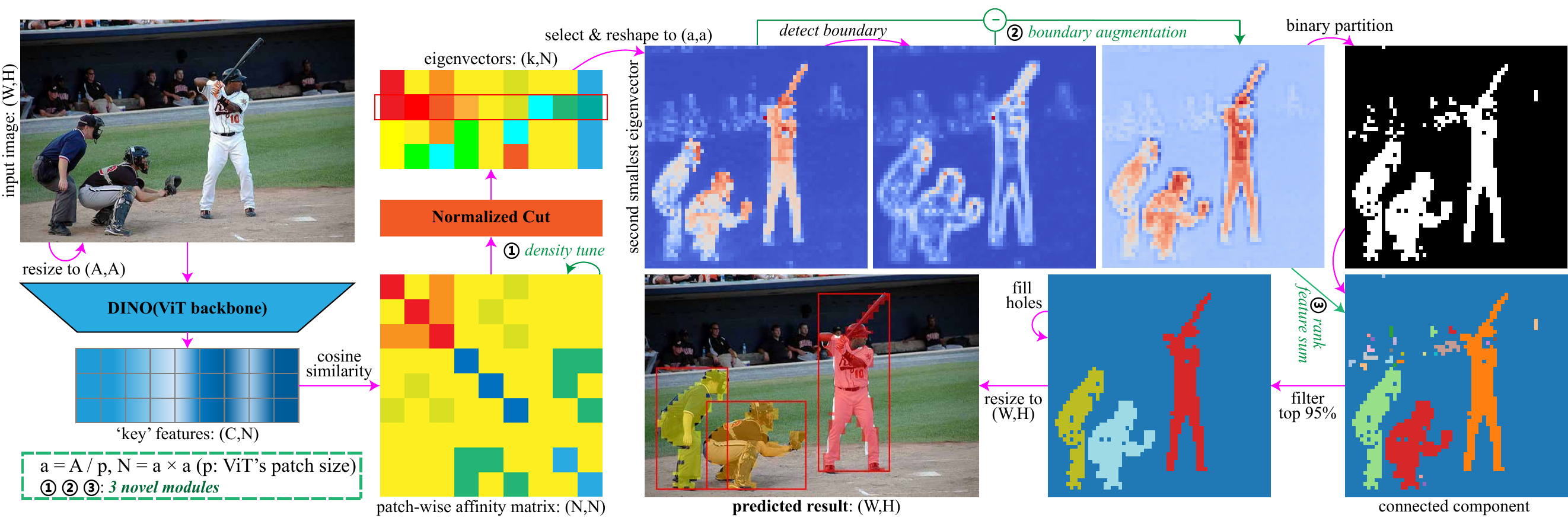}
	\caption{\textbf{Overview of CutOnce.} First, the resized image is processed by DINO to extract the ``key" features. Then, construct the affinity matrix and apply the NCut algorithm to obtain the second smallest eigenvector. Next, the original eigenvector is used to compute the boundary eigenvector, and the two are subtracted element-wise to produce the boundary-enhanced eigenvector. Finally, perform graph partitioning on the enhanced eigenvectors to generate segmentation masks.}
	\label{fig:cutonce_overview}
\end{figure}

\section{Method}
This chapter introduces a novel pipeline called \textbf{C}ut-\textbf{O}nce-and-\textbf{LE}a\textbf{R}n (COLER), designed for unsupervised instance segmentation and object detection.
We first propose CutOnce, a training-free method that efficiently generates masks for multiple objects. Built upon prior work, CutOnce enhances object discovery by optimizing the edge weighting scheme and introducing a boundary augmentation strategy in the NCut algorithm. In addition, it incorporates a rank feature filter to retain a sufficient number of valuable object masks.
Finally, we train a detector using the masks generated by CutOnce and further improve its performance through self-training.

\subsection{Preliminaries}
\myparagraph{Normalized Cut}
NCut~\citep{NCut} formulates image segmentation as a graph partitioning problem. It constructs a fully connected undirected graph $\mathbf{G} = (\mathbf{V}, \mathbf{E})$ by representing the image as a set of nodes, where each pair of nodes is connected by an edge with weight $w_{ij}$ indicating their similarity.
NCut minimizes the cost of partitioning the graph into two subgraphs by solving a generalized eigenvalue system
\begin{equation} \label{eq:eigenvector}
 (\mathbf{D} - \mathbf{W}) \mathbf{x} = \lambda \mathbf{D} \mathbf{x}
\end{equation}
to yield a set of $N \times N$ eigenvectors $\mathbf{x}$, where $N$ denotes the number of nodes. Here, $\mathbf{D}$ is an $N \times N$ diagonal matrix with $d_{ii} = \sum_{j} w_{ij}$, and $\mathbf{W}$ is an $N \times N$ symmetric matrix representing the adjacency matrix of edge weights.

\myparagraph{TokenCut and MaskCut}
Our CutOnce is \textit{based on the overall workflow of TokenCut}~\citep{TokenCut}, with some implementation details \textit{adopting the design of MaskCut}~\citep{CutLER}. The following describes the consistent parts of CutOnce with existing methods.

First, the input image is passed through a single self-supervised model to extract the ``key" features from the last attention layer, denoted as $\mathbf{K} \in \mathbb{R}^{D \times N}$, where $D$ is the feature dimension and $N$ is the number of nodes. The key feature of each patch is represented as a feature vector $\mathbf{k}_i$ ($i=1,\dots,N$). These features encode the spatial information captured by ViT~\citep{ViT}, so the cosine similarity between them can be used to calculate the elements in $\mathbf{W}$.
\begin{equation}
 w_{ij} = \cos(\mathbf{k}_i, \mathbf{k}_j) = \frac{\mathbf{k}_i^T \mathbf{k}_j}{\|\mathbf{k}_i\|_2 \, \|\mathbf{k}_j\|_2}
\end{equation}

Then, the second smallest eigenvector $\mathbf{y}_1$ is obtained from the solution of~\equref{eq:eigenvector}, which can be viewed as \textit{an enhanced attention map}.
When the splitting threshold is set to $\overline{\mathbf{y}_1} = \frac{1}{N} \sum_i \mathbf{y}_1^i$, $\mathbf{y}_1$ can be effectively divided into background and foreground.
To determine which group corresponds to the foreground, we examine the distribution of $\mathbf{y}_1$ on the ImageNet~\citep{ImageNet} and COCO~\citep{COCO} datasets, using criteria similar to MaskCut:
\textit{1)} The foreground set contains fewer than three of the four image corners, an idea inspired by the object-centric prior~\citep{BiasNCut}.
\textit{2)} $|\max(\mathbf{y}_1)| > |\min(\mathbf{y}_1)|$.
If either of these conditions is not satisfied (with condition 1 taking priority), the feature vector $\mathbf{v}$ used for mask partitioning is set to $-\mathbf{y}_1$; otherwise, it is set to $\mathbf{y}_1$. 
Finally, the groups in $\mathbf{v}$ with values greater than the partition point are regarded as foreground, while those with smaller values are regarded as background.

Existing methods have paved the way for our research, but the \textit{following limitations} remain urgent to address:
\textit{1)} Masks are not refined at the NCut stage.
\textit{2)} Predicted masks are prone to errors.
\textit{3)} Processing multiple targets is relatively time-consuming.

\subsection{CutOnce: Efficient Mask Generator}
Our CutOnce does not recurse NCut, does not rely on clustering, and does not use CRF to refine mask boundaries. \textit{It is, to our knowledge, the simplest multi-object mask generator to date, with computational complexity only slightly higher than TokenCut.}
Given the limitations of prior works, we aim to address the following challenges:
\textit{1)} How to produce more accurate foreground boundaries.
\textit{2)} How to enable NCut algorithm to discover multiple objects rather than focusing on a single one.
\textit{3)} How to segment more objects without introducing too many incorrect masks.

\myparagraph{Three Novel Modules in CutOnce}
The overall pipeline of CutOnce is illustrated in~\Figref{fig:cutonce_overview}.
The first two improvements optimize the \textit{input} and \textit{output} of the NCut algorithm, respectively. 
They all \textit{make the eigenvector distribution closer to the reality}, thereby addressing the first challenge.
The second improvement expands the foreground region, effectively resolving the second challenge.
The third improvement is applied in the graph partition stage, effectively filtering out the most salient multiple objects and addressing the third challenge.

\myparagraph{Density-Tune Cosine Similarity}
Previous methods~\citep{TokenCut,CutLER,CuVLER} compute the edge weight matrix $\mathbf{W}$ solely based on the cosine similarity between nodes, ignoring the variations in feature density across different image regions. This often leads to \textit{over-activation in certain areas}, which negatively affects boundary localization.
To address this issue, we propose a local-density-aware temperature modulation for cosine similarity. The idea is to adaptively adjust the temperature parameter in similarity computation based on the local density of feature points.

For the convenience of subsequent calculations, the deep learning features $\mathbf{K}$ are first normalized. The elements of the adaptive edge weight matrix are defined as:
\begin{equation}
 w_{ij} = \frac{\cos(\mathbf{k}_i, \mathbf{k}_j)}{T_{ij}} = \frac{\mathbf{k}_i^T \mathbf{k}_j}{T_0 + \alpha \cdot \frac{\rho_i + \rho_j}{2}}
\end{equation}
where $T_{ij}$ denotes the adaptive temperature parameter, $T_0$ is the base temperature, $\alpha$ is the modulation parameter, and $\rho_i$ and $\rho_j$ represent the local densities of feature points $i$ and $j$, respectively.
The local density is computed by first calculating the pairwise cosine similarity matrix $\mathbf{S} = \mathbf{K} \mathbf{K}^T$ for all patches in batch.
Then, for each feature point $i$, we select its top-$k$ most similar neighbors (excluding itself) and compute the local density as:
\begin{equation}
	\rho_i = \frac{1}{k} \sum_{j \in \mathcal{N}_k(i)} \mathbf{S}_{ij}
\end{equation}
where $\mathcal{N}k(i)$ denotes the set of indices corresponding to the $k$ most similar samples to the $i$-th sample (excluding $i$ itself).
The modulated $\mathbf{W}$ still requires feature contrast enhancement, following the approach in TokenCut~\citep{TokenCut}. 
Specifically, $\mathbf{W}_{ij}$ is set to 1 if $\mathbf{W}_{ij} \geq \tau^{\text{ncut}}$, and to $1e^{-5}$ otherwise, where $\tau^{\text{ncut}}$ is set to 0.15 by default.

Background regions typically exhibit relatively uniform features, allowing low-temperature areas to preserve the original similarity. In contrast, object interiors tend to have dense but uneven features, where high-temperature areas suppress similarity more strongly, making intra-object similarities more consistent. Obviously, regions with uniform similarity are more likely to be grouped together, and our improvement leads to more accurate separation of foreground and background.

This density-tune module shares the same idea as self-tuning spectral clustering~\citep{tuneSpectral}, \textit{both optimizing the affinity matrix via neighborhood information to reduce sensitivity}. Clearly, our approach is simpler and more intuitive.
A similar idea is adopted by the clustering algorithm LDP-SC~\citep{DensityPeakCluster}, which combines local density peaks with NCut and demonstrates significant advantages when handling locally tree-structured data. 

\myparagraph{Boundary Augmentation}
The attention map $\mathbf{y_1}$ output by NCut tends to focus on a single object, which is the fundamental reason why MaskCut~\citep{CutLER} applies the NCut algorithm multiple times to discover multiple objects. However, our goal is to obtain masks for multiple objects using NCut only once.
This raises an important question: Are the less salient objects in the foreground being ignored by the self-supervised model or the NCut algorithm?
In fact, the potential objects are already represented, but it is difficult to assign those with relatively low attention to the foreground.
From the first attention map (visualization of $\mathbf{y_1}$) in~\Figref{fig:cutonce_overview}, the following information can be easily extracted:
\textit{1)} Regions with larger feature values are usually concentrated within parts of the objects.
\textit{2)} In some object boundary areas, the feature values differ significantly from their surrounding regions.

To segment more objects, the salient regions within the foreground need to be expanded. Can boundary information be leveraged to achieve this? In practice, incorporating boundary information to refine original eigenvector $\mathbf{y_1}$ has proven to be an effective approach. We propose a boundary-enhanced feature representation:
\begin{equation} \label{eq:boundary}
	X_a = X - X_b
\end{equation}
where $X$ is the original eigenvector, and $X_b$ is the boundary eigenvector obtained by calculating the difference between each point and its neighborhood:
\begin{equation}
	X_b = \frac{1}{k} \sum_{n \in \mathcal{N}_k} |X - X_n|, \quad k \in \{4, 8\}
\end{equation}
Here, $\mathcal{N}_k$ denotes the $k$-neighborhood, which is set to 8 by default, and $X_n$ represents the feature values within the neighborhood. To correctly compute boundary pixels, padding is applied to the four edges and four corners of the feature map. The padding regions should not introduce new groups, so the original boundary features are simply extended. Specifically, the feature values in the padding areas are set to those of the adjacent boundary pixels.

The third attention map in \Figref{fig:cutonce_overview} demonstrates the benefits of this improvement: \textit{(1) The saliency of more objects is enhanced. (2) Nearby objects are less likely to be considered as a single entity.}
First, we analyze the first advantage.
$\mathbf{X}_b$ places high attention on regions with large feature differences, particularly on small areas around the boundary between foreground and background, as well as certain regions inside the foreground. After optimizing the attention distribution using~\equref{eq:boundary}, $\mathbf{X}_a$ exhibits \textit{generally reduced attention within the foreground} compared with $\mathbf{X}$, leading to smaller feature differences within the foreground. As a result, \textit{the salient regions of the foreground become larger}.
Next, we analyze the second advantage.
$\mathbf{X}_b$ shows high attention on both sides of object boundaries, which causes $\mathbf{X}_a$ to “merge” the areas around the boundary into the background, \textit{making all detected objects smaller}. For objects that are close to each other, the gaps between them are enlarged, which facilitates their correct separation. However, for small objects, prediction errors increase. If an object is very small in area, this may cause such “noisy points” to vanish.
Considering both comprehensive theoretical analysis and subsequent ablation study, boundary enhancement proves to be a strategy with more advantages than disadvantages. 
\textit{More visualizations of the computation process for this module are provided in the Appendix.}

The design of the boundary enhancement module is inspired by the residual connections in ResNet~\citep{ResNet}, where an auxiliary path is used to correct the main path.
Our $\mathbf{X}_b$ is a local difference (an approximation of a first-order gradient) computed from the eigenvector, \textit{capturing boundary-related variation patterns}. The idea is similar to the classical Laplacian of Gaussian (LoG)~\citep{edgeDetection} in image processing, \textit{where strong local changes (the sum of second-order derivatives) are used for edge detection}.

\myparagraph{Ranking-Based Instance Filter}
After extracting the foreground region from eigenvector using a segmentation threshold (as described in the \textit{Preliminaries} section), the next step is to separate multiple objects from the foreground. To achieve this, we first apply 4-connectivity to perform connected component decomposition on the foreground region, treating each connected component as a candidate object.
To select multiple salient objects from these candidates, we propose a feature rank-based connected component filtering strategy, which proceeds as follows:
\begin{enumerate}[leftmargin=2em, topsep=0pt]
\item Sort: Suppose there are $N$ candidate object regions. Let $s_i$ denote the feature sum of the $i$-th region. Sort all feature sums $\{s_i\}_{i=1}^N$ in descending order to obtain the index sequence $\{i_1, i_2, \ldots, i_N\}$.
\item Cumulative screening: Select the top-ranked objects one by one until the cumulative feature proportion reaches: $\frac{\sum_{j=1}^{k} s_{i_j}}{\sum_{i=1}^{N} s_i} \geq \tau$, where $\tau \in (0,1)$ is the feature preservation threshold.
\item Output: The masks corresponding to the top $k$ selected objects.
\end{enumerate}

Previous methods such as TokenCut and CutLER determine the sole target by checking the connected component containing the maximum absolute value in the eigenvector. This point-based criterion is prone to misidentification.
\textit{We instead use the features of a region rather than a single point to decide which object NCut prioritizes.} This approach accounts for both the spatial extent of an object and the feature magnitude at individual points, allowing us to select the most salient objects and output them in descending order.
More importantly, it introduces only a single hyperparameter.

\subsection{Self-Training.}
After training the detector with pseudo labels generated by CutOnce, the detector can identify more masks than those in the pseudo labels. Therefore, we adopt a self-training strategy to further improve the model performance.
In the $t$-th round of self-training ($t \in {1, 2, \ldots}$), we first perform inference on the training data using the current model, retaining predicted masks with confidence scores higher than $0.6 - 0.05t$ as high-quality pseudo-labels. To avoid label duplication and maintain diversity in the training data, we also select a subset of pseudo-labels from round $(t{-}1)$ whose IoU with the current high-confidence predictions is $<$ 0.5. The final training labels for round $t$ are obtained by merging these two sets.

\section{Experiments}
\label{sec:experiments}

\subsection{Implementation Details}
\myparagraph{Datasets}
In this paper, only the images from the train split of ImageNet-1K~\citep{ImageNet} (1.28M images) are used for all training processes of the COLER model, with no manual annotations or any supervised pre-trained models employed in the training.
\textit{Due to limited computational resources, we use the ImageNet val split (50K images) to generate pseudo-labels in our ablation study, while keeping all other settings identical.} This variant is denoted as CutOnce* and COLER*.

We evaluate on two subsets of the COCO~\citep{COCO} dataset, LVIS~\citep{LVIS}, VOC~\citep{VOC}, KITTI~\citep{KITTI}, OpenImages~\citep{OpenImages} and Objects365~\citep{Objects365}, resulting in a total of 7 benchmarks.
This paper mainly uses AP$_{50}$ and AP as the evaluation metrics for presenting results.

\myparagraph{CutOnce} 
We resize images to 480$\times$480 pixels and use the ViT-B/8~\citep{ViT} DINO~\citep{DINO} model by default to extract features. For the density-tune similarity module, $k$, $T_0$, and $\alpha$ are set to 10, 1.0, and 0.5, respectively. 
The filter parameter $\tau$ is set to 0.95.

\myparagraph{CAD (class-agnostic detector)}
\textit{All training and inference are conducted on single NVIDIA RTX 4090 GPU.} All experiments are implemented on the detectron2~\citep{detectron2} platform using Cascade Mask R-CNN~\citep{CascadeRCNN} as the default detector. The detector is trained with masks and bounding boxes generated by CutOnce for 80K iterations with copy-paste augmentation~\citep{copypaste}. The batch size is set to 8, learning rate to 0.01, weight decay to $5 \times 10^{-5}$, and momentum to 0.9. Following the same setup as the copy-paste augmentation in CutLER~\citep{CutLER}, we randomly downsample the masks with a scale factor uniformly sampled from 0.3 to 1.0.

\myparagraph{Self-Training}
In this stage, the model is initialized with the weights from the previous phase and trained for 60K iterations. Other settings follow CutLER, with the learning rate set to 0.005 and the copy-paste augmentation scalar uniformly sampled between 0.5 and 1.0.

\myparagraph{State-of-the-art (SOTA) Comparison}
We compare our method with CutLER~\citep{CutLER} and CuVLER~\citep{CuVLER}.
CuVLER provides two versions of pretrained weights: zero-shot and COCO self-train, and we only use the former.
\textit{Some methods are excluded from comparison.} For example, CutS3D~\citep{CutS3D} has not released its source code and weights, and unMORE leverages the COCO dataset for training, which is incompatible with zero-shot evaluation.

\begin{table}[t]
	\centering
	\tablestyle{0.6pt}{1.0}
	\small
	\caption{\textbf{Evaluation of pseudo labels.} \#N denotes the average number of masks per image.}
	\begin{tabular}{lccccc|cccccccccc|c}
		\toprule
		Datasets $\rightarrow$ & \multirow{2}{*}{\begin{tabular}[c]{@{}c@{}}Use~~~\\CRF~~~\end{tabular}} &
		\multicolumn{5}{c}{ImageNet val} &~& \multicolumn{9}{c}{COCO val2017} \\
		\cline{3-7} \cline{9-17}
		Methods && AP$^\text{box}$ & AP$_{50}^\text{box}$ & AP$_{75}^\text{box}$ & AR$_{100}^\text{box}$ & \#N && AP$^\text{box}$ & AP$_{50}^\text{box}$ & AP$_{75}^\text{box}$ & AR$_{100}^\text{box}$ & AP$^\text{mask}$ & AP$_{50}^\text{mask}$ & AP$_{75}^\text{mask}$ & AR$_{100}^\text{mask}$ & \#N \\
		\midrule
		MaskCut & \checkmark & 10.6 & 20.3 &10.0 &27.7 & 1.9 && 3.9 & 7.9 & 3.3 & 7.7 & 3.1 & 6.8 & 2.5 & 6.5 & 1.9 \\
		VoteCut & \checkmark & \textbf{20.9} & \textbf{36.2} & \textbf{20.0} & \textbf{45.0} & 8.9 && \textbf{5.5} & \textbf{10.7} & \textbf{4.9} & \textbf{12.2} & \textbf{4.5} & \textbf{9.3} & \textbf{3.9} & \textbf{10.3} & 8.6 \\
		CutOnce(ours) & \texttimes & 16.5 & 32.5 & 15.0 & 31.5 & 1.8 && 4.1 & 8.2 & 3.6 & 7.6 & 3.1 & 7.0 & 2.4 & 6.0 & 1.8 \\
		CutOnce+(ours) & \checkmark & 16.9 & 32.6 & 15.4 & 32.3 & 1.8 && 4.2 & 8.2 & 3.7 & 7.9 & 3.4 & 7.2 & 2.9 & 6.8 & 1.8 \\
		\bottomrule
	\end{tabular}
	\label{tab:cutonce}
\end{table}

\subsection{Pseudo Labels Evaluation}
\label{sec:pseudo_eval}

\begin{wrapfigure}{r}{0.6\textwidth}
	\vspace{-1.0em}
	\includegraphics[width=1.0\linewidth]{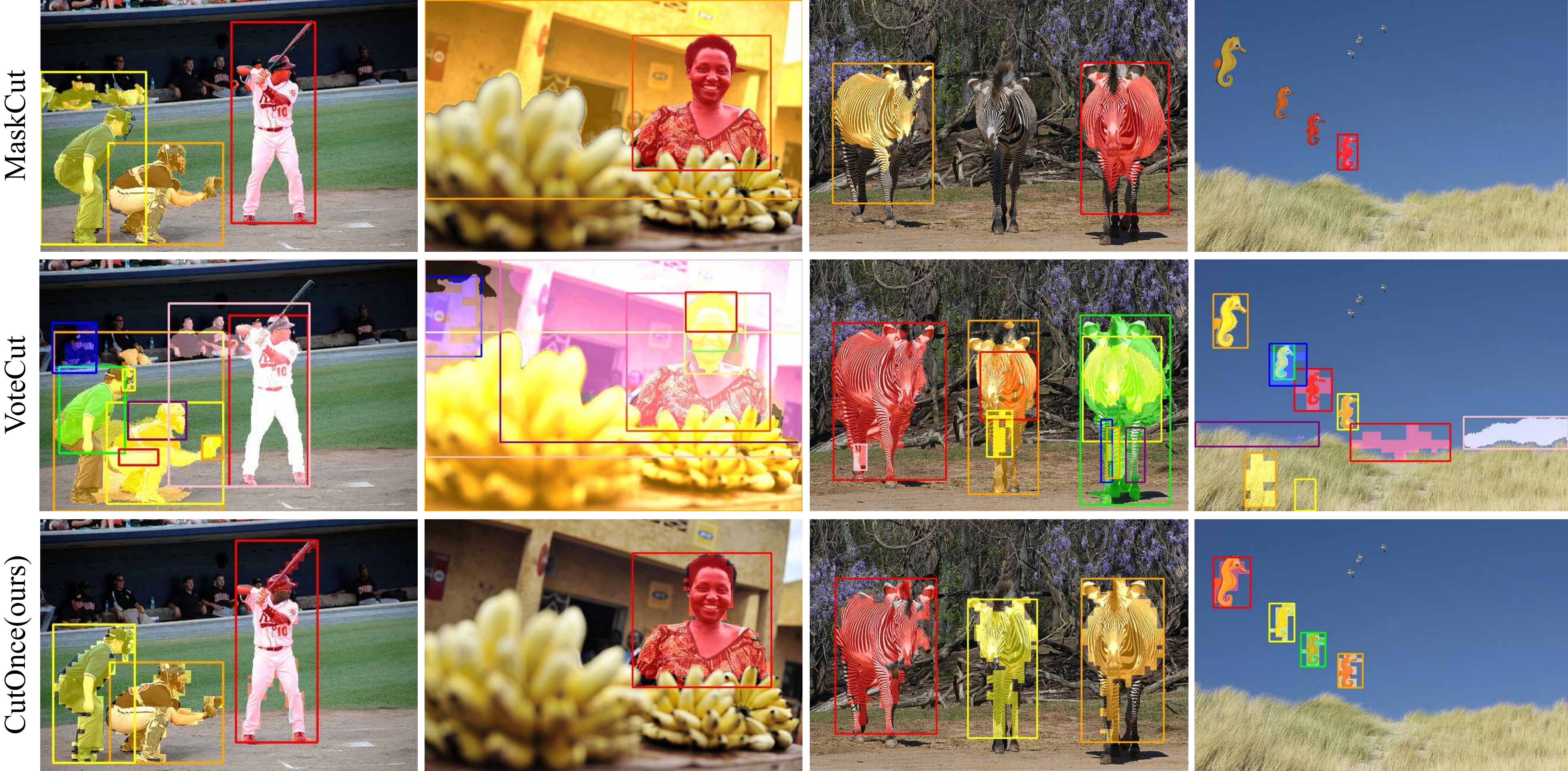}
	\caption{\textbf{Qualitative comparison between our CutOnce and related methods on COCO val2017.} All methods display all predicted masks.
		}
	\label{fig:cutonce_compare}
\end{wrapfigure}
To evaluate the pseudo masks using the official COCO~\citep{COCO} evaluation tool, each annotation must be assigned a confidence score. \textit{The score has a significant impact on AP but does not affect AR.} Since VoteCut~\citep{CuVLER} comes with its own scoring mechanism, we retain its original setting. To ensure a relatively fair comparison, we apply the same scoring scheme to both MaskCut~\citep{CutLER} and CutOnce.
The reason is that both methods output masks in descending order of object saliency and treat all outputs as ``ground truth".
For multiple masks associated with the same image, we adopt a linearly decreasing score assignment scheme.
The confidence score for each mask is defined as 1.0 if $N=1$, and as $1.0 - \tfrac{k}{2N-2}$ if $N>1$, where $N$ is the total number of masks in the image and $k$ is the index of the current mask ($k = 0, 1, \ldots, N-1$).
This ensures that the first mask always receives the highest confidence score of 1.0, while the last mask is assigned a score of 0.5.

\tabref{tab:cutonce} presents the quantitative comparisons of different methods, and \Figref{fig:cutonce_compare} shows the corresponding qualitative results. On ImageNet val, all metrics of CutOnce lie between those of MaskCut and VoteCut. On COCO val2017, CutOnce shows poor performance in AR$_{100}^\text{mask}$. 
From the performance of CutOnce+ (The only difference from CutOnce is whether CRF is used or not), it clearly outperforms CutOnce and MaskCut, achieving substantial improvements in all metrics except AP$_{50}$. Evidently, using CRF to refine the masks has a positive effect on high-precision localization, while having little impact on coarse localization, ultimately leading to significant gains in AP and AR.
As shown in \Figref{fig:cutonce_compare}, the targets localized by CutOnce are always correct. \textit{Since CutOnce does not use CRF for post-processing, its boundaries are slightly coarse compared to other methods.}
MaskCut applies NCut three times and can localize at most three objects. This design clearly cannot adapt to scenarios with either many objects or very few.
VoteCut uses clustering and integrates outputs from multiple self-supervised models, producing a large number of masks, some correct and some not meaningful.
Although VoteCut appears to achieve the highest metrics, \textit{its incorrect masks can negatively impact training}.
Overall, the pseudo labels generated by CutOnce contain relatively fewer noisy labels, which is beneficial for model learning.

\begin{table}[t]
	\centering
	\tablestyle{0.15pt}{1.0}
	\small
	\caption{
		\textbf{Zero-shot evaluation across three COCO-based datasets.}
		IN and `1 + 3' denote ImageNet-1K and one training plus three rounds of in-domain self-training, respectively.}
	\begin{tabular}{lccccccccccccccccc} 
		\toprule
		Datasets $\rightarrow$ & \multirow{2}{*}{Pretrain} & \multirow{2}{*}{\begin{tabular}[c]{@{}c@{}}Train\\ \#Rounds\end{tabular}} && \multicolumn{4}{c}{COCO 20K} &  & \multicolumn{4}{c}{COCO val2017} &  & \multicolumn{4}{c}{LVIS} \\ 
		\cline{5-8}\cline{10-13}\cline{15-18}
		Methods &  &  && AP$^\text{box}$ & AP$_{50}^\text{box}$ & AP$^\text{mask}$ & AP$_{50}^\text{mask}$ &  & AP$^\text{box}$ & AP$_{50}^\text{box}$ & AP$^\text{mask}$ & AP$_{50}^\text{mask}$ &  & AP$^\text{box}$ & AP$_{50}^\text{box}$ & AP$^\text{mask}$ & AP$_{50}^\text{mask}$ \\ 
		\midrule
		CutLER & IN train & 1 + 3 && 12.5 & 22.4 & 10.0 & 19.6 &  & 12.3 & 21.9 & 9.7 & 18.9 &  & 4.5 & 8.4 & 3.5 & 6.7 \\
		CuVLER & IN val & 1 && 12.7 & 23.5 & 10.0 & 20.1 &  & 12.6 & 23.0 & 9.8 & 19.3 &  & 4.5 & 8.6 & 3.6 & 6.9 \\
		COLER(ours) & IN train & 1 + 3 && 13.3 & 25.2 & 10.8 & 22.3 &  & 13.1 & 24.9 & 10.5 & 22.1 &  & 5.0 & 9.6 & 4.0 & 8.1 \\
		COLER*(ours) & IN val & 1 + 1 && 12.6 & 24.1 & 9.8 & 20.5 &  & 12.5 & 23.8 & 9.6 & 20.1 &  & 4.6 & 9.2 & 3.7 & 7.3 \\
		\midrule
		\textit{vs. SOTA} (\%) & & && \green{4.7} & \green{7.2} & \green{8.0} & \green{10.9} && \green{4.0} & \green{8.3} & \green{7.1} & \green{14.5} && \green{11.1} & \green{11.6} & \green{11.1} & \green{17.4} \\
		\bottomrule
	\end{tabular}
	\label{tab:zero_shot_coco}
\end{table}

\begin{table}[t]
	\centering
	\tablestyle{0.1pt}{1.0}
	\caption{
		\textbf{Zero-shot unsupervised object detection evaluation.}
		Avg. denotes the average value.}
	\begin{tabular}{l|cc|cc|cc|cc|cc|cc|cc|cc} 
		\toprule
		Datasets $\rightarrow$     & \multicolumn{2}{c|}{Avg.} & \multicolumn{2}{c|}{COCO} & \multicolumn{2}{c|}{COCO20K} & \multicolumn{2}{c|}{LVIS} & \multicolumn{2}{c|}{VOC} & \multicolumn{2}{c|}{KITTI} & \multicolumn{2}{c|}{OpenImages} & \multicolumn{2}{c}{Objects365} \\
		Metrics $\rightarrow$     & AP$_{50}$ & AP         & AP$_{50}$ & AP         & AP$_{50}$ & AP   & AP$_{50}$ & AP         & AP$_{50}$ & AP        & AP$_{50}$ & AP   & AP$_{50}$ & AP   & AP$_{50}$ & AP \\ 
		\midrule
		CutLER & 21.0 & 11.3 & 21.9 & 12.3 & 22.4 & 12.5 & 8.4 & 4.5 & 36.9 & 20.2 & 18.4 & 8.5 & 17.3 & 9.7 & 21.6 & 11.4 \\
		CuVLER & 21.2 & 11.4 & 23.0 & 12.6 & 23.5 & 12.7 & 8.6 & 4.5 & 39.4 & 22.3 & 13.0 & 5.1 & 19.6 & 11.6 & 21.6 & 10.9	\\
		COLER(ours) & 23.7 & 12.1 & 24.9 & 13.1 & 25.2 & 13.3 & 9.6 & 5.0 & 41.8 & 21.6 & 22.3 & 9.6 & 18.2 & 10.1 & 23.9 & 12.0 \\
		COLER*(ours) & 22.3 & 11.4 & 23.8 & 12.5 & 24.1 & 12.6 & 9.2 & 4.6 & 39.1 & 20.5 & 20.8 & 8.8 & 16.7 & 9.3 & 22.6 & 11.2	\\
		\midrule
		\textit{vs. SOTA} (\%) & \green{11.6} & \green{6.3} & \green{8.3} & \green{4.0} & \green{7.2} & \green{4.7} & \green{11.6} & \green{11.1} & \green{6.1} & -3.1 & \green{21.2} & \green{12.9} & -7.1 & -12.9 & \green{10.6} & \green{5.3} \\
		\bottomrule
	\end{tabular}
	\label{tab:zero_shot_detection}
\end{table}

\subsection{Unsupervised Zero-shot Evaluations}
We evaluate COLER on \textit{7 different benchmarks}, containing a variety of object categories and image styles, to validate its effectiveness as a general unsupervised method. 

\begin{wrapfigure}{r}{0.5\textwidth}
    \centering
    \includegraphics[width=1.0\linewidth]{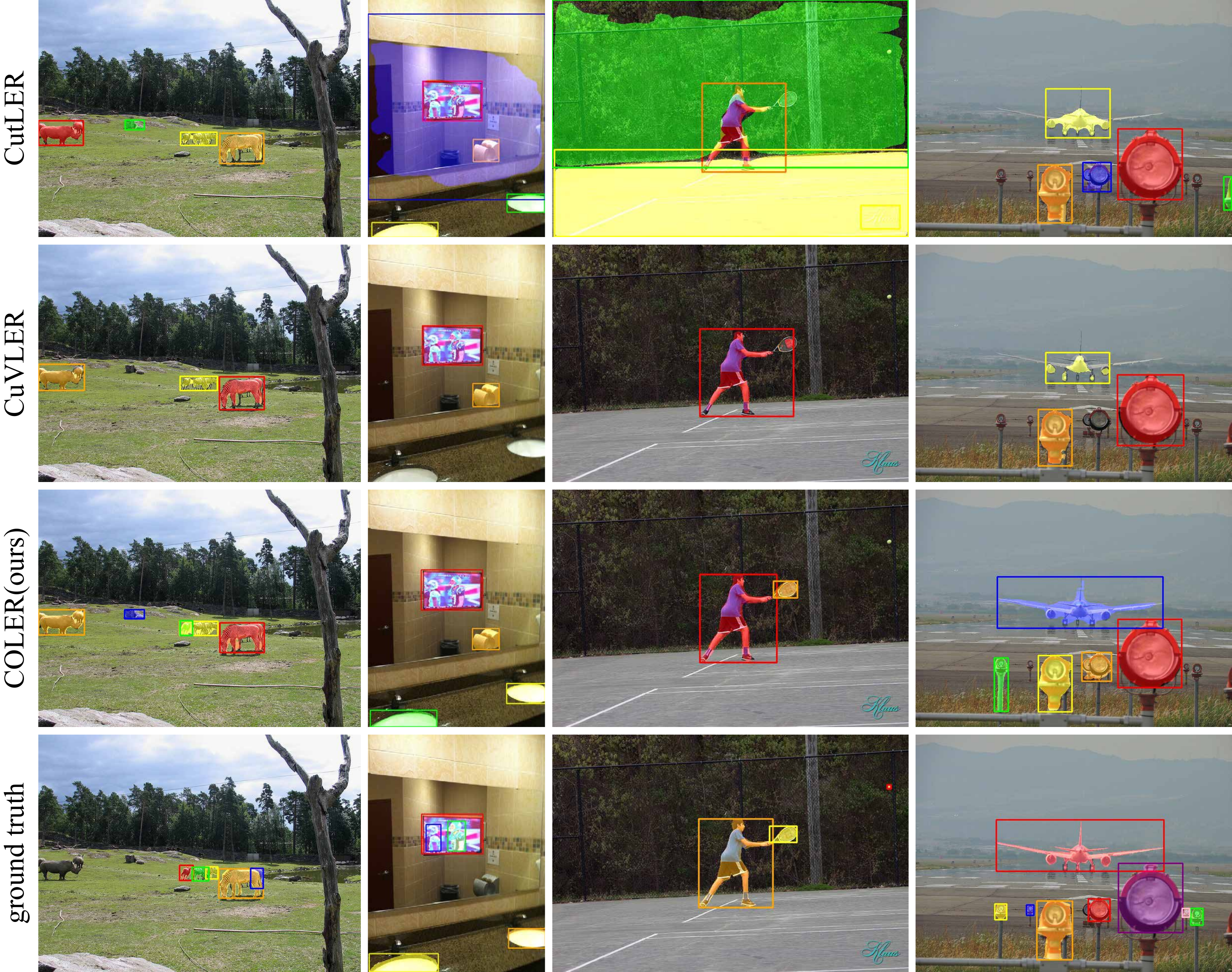}
    \caption{\textbf{Qualitative comparison between our COLER and SOTA methods on COCO val2017.}
 Only predictions with confidence $\geq$ 0.5 are shown.
 }
    \label{fig:coler_compare}
\end{wrapfigure}
\myparagraph{Detailed in COCO Datasets}
COCO 20K, COCO val2017, and LVIS are all from COCO and provide segmentation annotations, with the corresponding results reported in \tabref{tab:zero_shot_coco}. Our COLER achieves leading performance on all three datasets, and COLER* performs comparably to prior methods. The number of self-training iterations reported in \tabref{tab:zero_shot_coco} corresponds to the maximum iterations that continue to improve performance. Both CutLER and our COLER stop improving after three rounds of self-training. \textit{CuVLER does not provide in-domain self-trained weights}, and its source code does not get weights better than the zero-shot version. Our COLER* stops improving after the second round of self-training. In terms of percentage improvement over SOTA, COLER achieves larger gains on the densely annotated LVIS dataset.
\Figref{fig:coler_compare} shows the qualitative results of COLER compared with related methods. Obviously, \textit{COLER often detects more useful instances, including some that are not annotated in the ground truth}.

\myparagraph{Object Detection}
In \tabref{tab:zero_shot_detection}, we report COLER's object detection performance across all datasets. 
On average, COLER shows a larger improvement in AP${50}$ but a smaller gain in AP. This is expected, as the absence of CRF makes high-precision localization less evident than the improvement seen in AP${50}$.
Comparing the best results on each dataset, COLER shows significant advantages on KITTI and LVIS, while performing the worst on OpenImages. 
The prediction performance of COLER* is also acceptable, which is consistent with the experimental results in the previous section. Overall, our method achieves certain advantages across all datasets.

\begin{table}[t]
	\centering
	\tablestyle{1pt}{1.0}
	\begin{tabular}{lccgcllccgcllcgcllcgc}
		\cmidrule[\heavyrulewidth]{1-5}\cmidrule[\heavyrulewidth]{7-11}\cmidrule[\heavyrulewidth]{13-16}\cmidrule[\heavyrulewidth]{18-21}
		$\tau$  & 0.8 & 0.9 & 0.95 & 0.99 && $k$ & 3 & 5 & 10 & 20 && $T_0$ & 0.8 & 1.0 & 1.2 && $\alpha$ & 0.3 & 0.5 & 0.7 \\ 
		\cmidrule{1-5}\cmidrule{7-11}\cmidrule{13-16}\cmidrule{18-21}
		AP$^\text{mask}_{50}$ & 17.7 & 18.9 & 19.6 & 17.5 &&  AP$^\text{mask}_{50}$ & 18.9 & 19.3 & 19.6 & 18.7 && AP$^\text{mask}_{50}$ & 18.5 & 19.6 & 18.9  && 
		AP$^\text{mask}_{50}$& 18.7 & 19.6 & 19.0 \\
		\cmidrule[\heavyrulewidth]{1-5}\cmidrule[\heavyrulewidth]{7-11}\cmidrule[\heavyrulewidth]{13-16}\cmidrule[\heavyrulewidth]{18-21}
	\end{tabular}
	\caption{\textbf{Ablation study of CutOnce* hyperparameters on COCO val2017 .} $\tau$ denotes the preservation ratio in the filter, while $k$, $T_0$, and $\alpha$ are parameters related to the adaptive edge weight matrix.}
	\label{tab:ablation_cutonce}
\end{table}

\begin{figure}[t]
	\centering
	\includegraphics[width=1.0\linewidth]{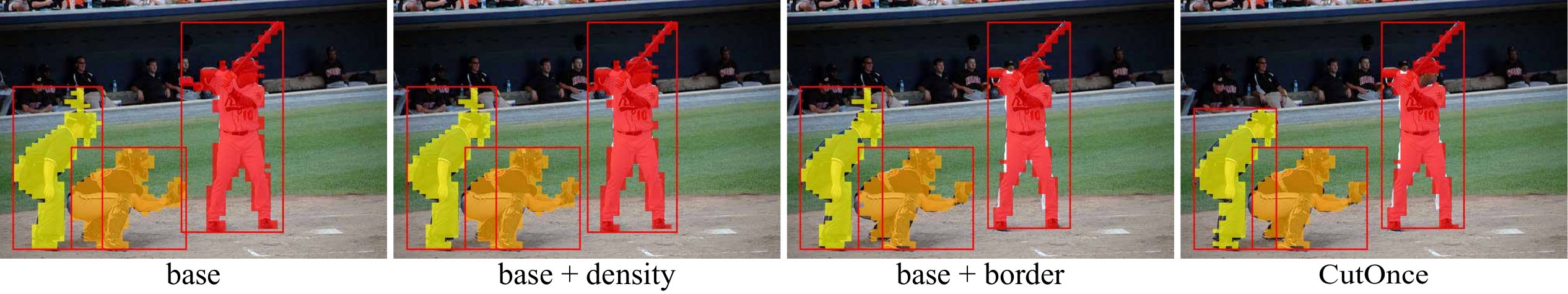}
	\caption{\textbf{Ablation study on the two NCut refinement modules in CutOnce.} 
	}
	\label{fig:ablation_cutonce}
\end{figure}

\subsection{Ablation Study}
\tabref{tab:ablation_cutonce} reports the \textit{ablation study of the hyperparameters introduced by CutOnce*, where the detector is trained only once.}

\Figref{fig:ablation_cutonce} visualizes the ablation study of two NCut refinement modules introduced in CutOnce for enhanced object discovery. It is evident that when the two modules are combined, CutOnce can finely segment the three most salient targets, whereas removing either module often results in poor mask boundaries for certain objects. 

\begin{wraptable}{r}{0.53\textwidth}
	\vspace{-1.0em}
	\captionof{table}{\textbf{Ablation analysis of COLER components on COCO val2017 and KITTI.}}
	\tablestyle{0pt}{1.0}
	\begin{tabular}{lccccc}
		\toprule
		\multirow{2}{*}{Methods} & \multicolumn{2}{c}{COCO} &\kern0.5em& \multicolumn{2}{c}{KITTI} \\
		\cline{2-3} \cline{5-6}
		& AP$^\text{mask}$ & AP$^\text{mask}_{50}$ && AP$^\text{box}$ & AP$^\text{box}_{50}$ \\
		\midrule
		TokenCut + CAD & 6.2 & 14 && 5.7 & 14.1 \\
		\midrule
		+ rank feature filter & 8.1 & 16.1 && 6.7 & 15.8 \\
		+ similarity tune & 8.4 & 16.5 && 7.1 & 16.4 \\
		+ boundary augment & 9.1 & 19.1 && 8.2 & 19.2 \\
		\midrule
		+ copy-paste & 9.6 & 20.6 && 8.6 & 20.1 \\ \rowcolor{gray!30}
		+ self-training (COLER) & 10.5 & 22.1 && 9.6 & 22.3 \\
		\bottomrule
	\end{tabular}
	\label{tab:ablation_coler}
	\vspace{-1.0em}
\end{wraptable}
\tabref{tab:ablation_coler} reports the impact of each component in COLER on the final results across two datasets. Each component contributes to performance improvements to varying degrees, but boundary augmentation module proves to be the most critical for boosting performance.
Additionally, copy-paste augmentation~\citep{copypaste} and self-training during the training process also contribute significantly to the overall improvement of COLER.

\tabref{tab:self_train} reports the impact of the number of self-training rounds on the final results of COLER. 
The results after the first, second, and third rounds of self-training show steady improvements, while the gain in the fourth round becomes almost negligible.
\textit{By default, COLER uses 3 rounds of self-training.}

\subsection{Discussion}
\begin{wraptable}{r}{0.47\textwidth}
	\vspace{-1.em}
	\tablestyle{0pt}{1.0}
	\captionof{table}{\textbf{Number of self-training rounds in COLER.}}
	\begin{tabular}{lcccclcc} 
		\toprule
		\multirow{2}{*}{Round } & \multicolumn{4}{c}{COCO} &\kern0.5em& \multicolumn{2}{c}{KITTI} \\
		\cline{2-5} \cline{7-8}
		& AP$^\text{box}$ & AP$^\text{box}_{50}$ & AP$^\text{mask}$ & AP$^\text{mask}_{50}$ && AP$^\text{box}$ & AP$^\text{box}_{50}$ \\
		\midrule
		0 & 12.6 & 23.9 & 9.6 & 20.1 && 8.5 & 19.9 \\
		1 & 12.7 & 24.2 & 9.9 & 20.7 && 8.8 & 20.8 \\
		2 & 13.0 & 24.7 & 10.3 & 21.6 && 9.4 & 21.8 \\ \rowcolor{gray!30}
		3 & 13.1 & 24.9 & 10.5 & 22.1 && 9.6 & 22.3 \\
		4 & 13.1 & 24.9 & 10.5 & 22.3 && 9.6 & 22.4 \\
		\bottomrule
	\end{tabular}
	\label{tab:self_train}
	\vspace{-1.em}
\end{wraptable}

Despite the strong performance of COLER, it also has several limitations. Examples of failure cases are shown in \figref{fig:coler_failure}:
\textit{1)} For heavily overlapping objects with unclear boundaries, the ability to distinguish them correctly is limited.
\textit{2)} The number of detected objects is also constrained; although CutOnce can detect more than ten objects, it ultimately depends on the object discovery capability of the self-supervised model.
\textit{3)} Its understanding of dense scenes remains weak, even though it shows notable improvements over previous methods.

\begin{figure}[t]
	\centering
	\includegraphics[width=1.0\linewidth]{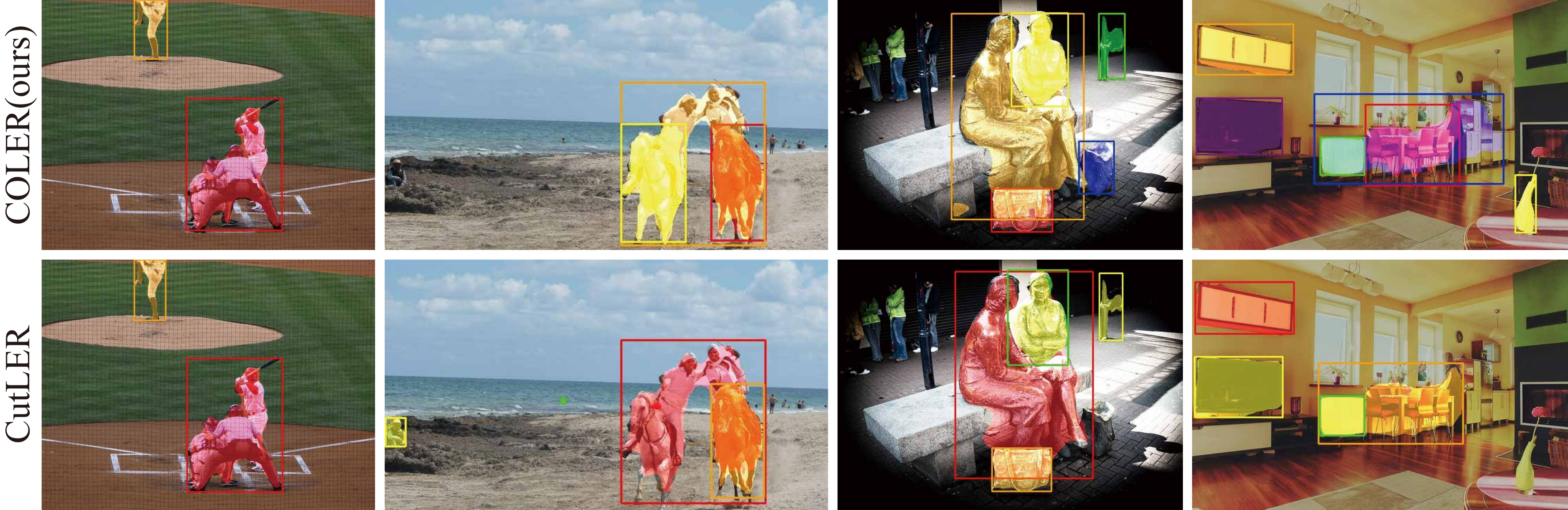}
	\caption{\textbf{Failure cases of COLER on COCO val2017.}}
	\label{fig:coler_failure}
\end{figure}

\section{Conclusion}
We propose CutOnce, a novel training-free method for unsupervised object discovery that efficiently and accurately partitions multiple instances within single image. In particular, the \textit{boundary augmentation strategy stands out as the simplest yet most effective improvement} in this work, and we believe it holds great potential for broader applications in the future.
We also introduce COLER, a zero-shot model trained using masks generated by CutOnce. With only ImageNet-1K as the source domain, COLER surpasses previous state-of-the-art models across multiple benchmarks in both unsupervised instance segmentation and object detection.

\bibliography{iclr2026}

\bibliographystyle{iclr2026_conference}

\input{appendix}
\fi

\end{document}

%% file: math_commands.tex
\usepackage{amsmath,amsfonts,bm}

\def\figref#1{figure~\ref{#1}}
\def\Figref#1{Figure~\ref{#1}}

\def\eqref#1{equation~\ref{#1}}

\def\1{\bm{1}}

\DeclareMathAlphabet{\mathsfit}{\encodingdefault}{\sfdefault}{m}{sl}
\SetMathAlphabet{\mathsfit}{bold}{\encodingdefault}{\sfdefault}{bx}{n}



%% file: appendix.tex
\clearpage

\appendix
\section{Appendix}

\subsection{Datasets used for evaluation}
\noindent\textbf{COCO}~\cite{COCO} (Microsoft Common Objects in Context) is a large-scale dataset for object detection and segmentation. In this paper, COCO refers to the 5k images from the \texttt{2017 validation} set.

\begin{wraptable}{r}{0.6\textwidth}
	\vspace{-1.0em}
	\centering
	\setlength\tabcolsep{1pt}
	\caption{\textbf{Summary of datasets used for zero-shot evaluation (except ImageNet).} "avg. \# obj." denotes the average number of annotations per image.}
	\begin{tabular}{lcccc}
		\toprule
		datasets      & testing data   & seg label & \#images & avg. \# obj. \\ 
		\midrule
		COCO  & val2017        & \checkmark & 5,000   & 7.4 \\
		COCO20K       & train2014 & \checkmark & 19,817  & 7.3 \\
		LVIS  & val    & \checkmark & 19,809  & 12.4        \\
		Pascal VOC    & trainval07     & \texttimes & 9,963   & 3.1 \\
		KITTI & trainval    & \texttimes        & 7,521  & 4.7 \\
		OpenImages V7 & val    & \texttimes        & 41,620  & 7.3 \\
		Object365 V2 & val    & \texttimes        & 80,000  & 15.5 \\ 
		\midrule
		ImageNet      & val    & \texttimes & 50,000  & 1.6 \\ 
		\bottomrule
	\end{tabular}
	\label{tab:datasets_zero_shot}
	\vspace{-1.0em}
\end{wraptable}
\noindent\textbf{COCO 20K}\cite{COCO} contains 19,817 images, a subset of COCO train2014. Many previous unsupervised methods\cite{TokenCut,CutLER,CuVLER} have used this dataset to evaluate model performance.

\noindent\textbf{LVIS}~\cite{LVIS}: (Large Vocabulary Instance Segmentation) is a dataset for long-tail instance segmentation. It contains 2.2 million high-quality instance masks of over 1,000 entry-level object categories, collected based on the COCO dataset. In this paper, LVIS refers to the 19,809 images in the \texttt{validation} set.

\noindent\textbf{VOC}~\cite{VOC} (PASCAL Visual Object Classes) is a widely used benchmark for object detection. We evaluate on its \texttt{trainval07} split.

\noindent\textbf{KITTI}~\cite{KITTI} (Karlsruhe Institute of Technology and Toyota Technological Institute) is one of the most popular datasets for mobile robotics and autonomous driving. We evaluate on its \texttt{trainval} split.

\noindent\textbf{OpenImages V7}~\cite{OpenImages} contains multiple tasks, including image classification, object detection, instance segmentation, and visual relationship detection. We evaluate on over 40K images from the \texttt{val} split.

\noindent\textbf{Object365 V2}~\cite{Objects365} provides a supervised object detection benchmark with a focus on diverse objects in the natural world. We evaluate on 80K images from the \texttt{val} split.

The summary of these datasets used for zero-shot evaluation is provided in \tabref{tab:datasets_zero_shot}.


\begin{figure}[h]
	\centering
	\includegraphics[width=1.0\linewidth]{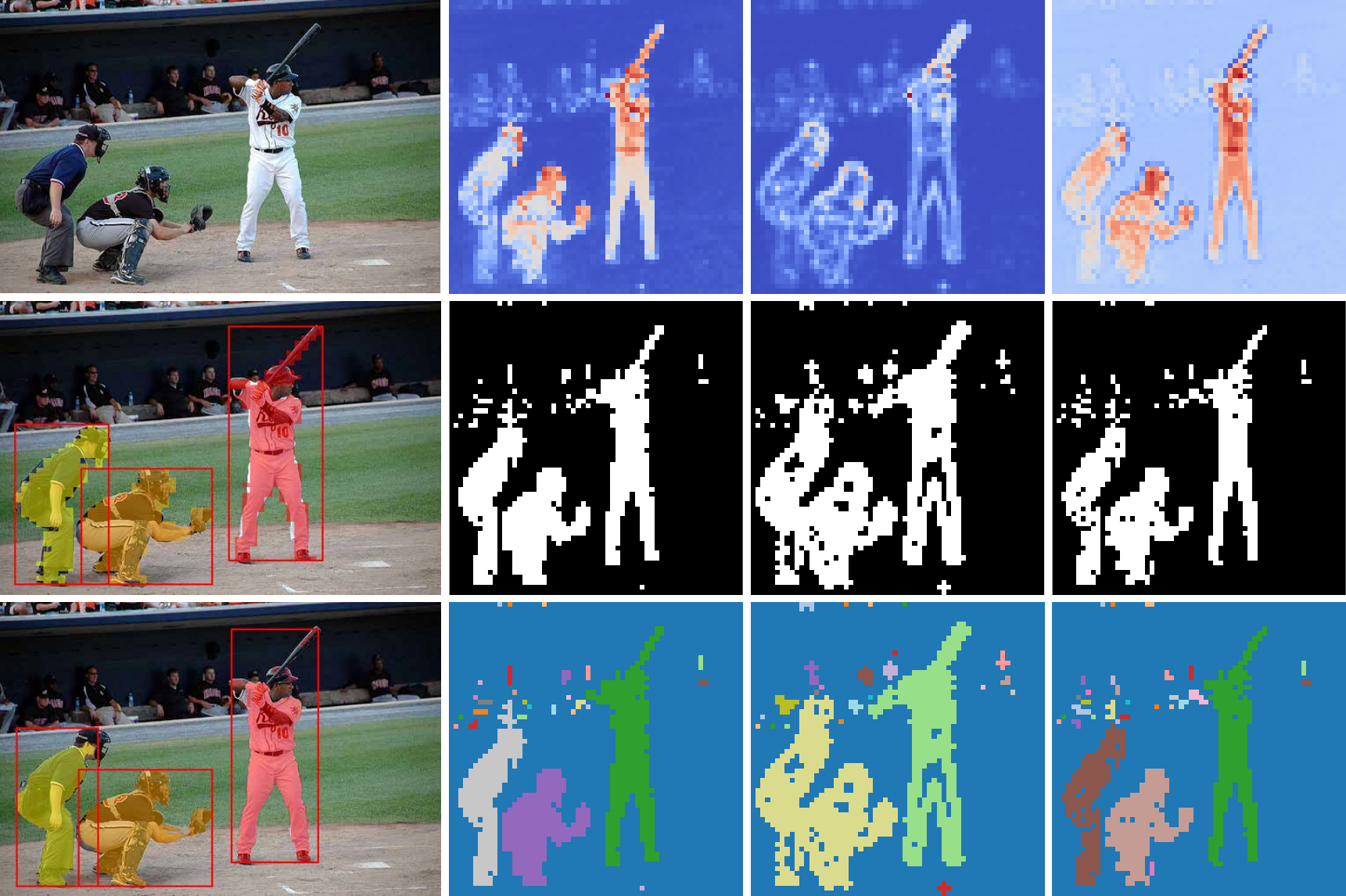}
	\caption{\textbf{Visualization of the computation process of CutOnce.} The first column (top to bottom) shows the original image, CutOnce prediction, and CutOnce with post-processing. The second to fourth columns show \textit{raw eigenvector}, \textit{boundary eigenvector}, \textit{difference between the two}, and the corresponding foreground-background \textit{binary maps} and \textit{connected component maps}.}
	\label{fig:cutonce_details}
\end{figure}

\begin{figure}[h]
	\centering
	\includegraphics[width=1.0\linewidth]{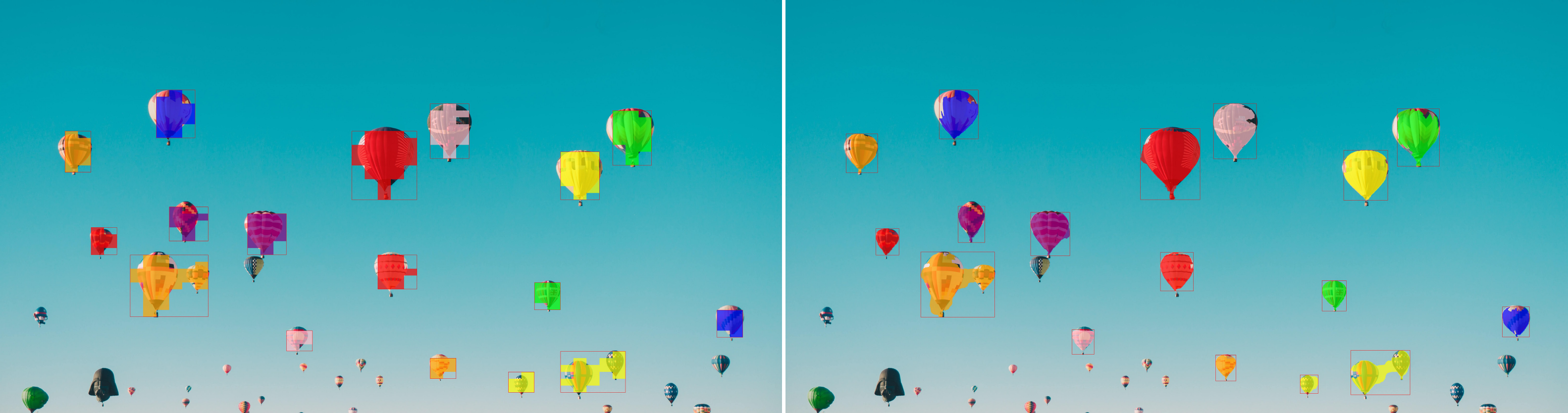}
	\caption{\textbf{Detecting many objects with CutOnce}. The first row shows the results of CutOnce, and the second row presents the results of CutOnce with post-processing. Both methods successfully detect \textit{17 objects}.}
	\label{fig:more_objects_cutonce}
\end{figure}

\begin{table}[h]
	\centering
	\tablestyle{0pt}{1.0}
	\caption{\textbf{Unsupervised instance segmentation results on all benchmarks in this work.}}
	\begin{tabular}{l|cccccc|cccccc} 
		\toprule
		Datasets & AP$^\text{mask}$ & AP$_{50}^\text{mask}$ & AP$_{75}^\text{mask}$ & AP$_{S}^\text{mask}$ & AP$_{M}^\text{mask}$ & AP$_{L}^\text{mask}$ & AR$_{1}^\text{mask}$ & AR$_{10}^\text{mask}$ & AR$_{100}^\text{mask}$ & AR$_{S}^\text{mask}$ & AR$_{M}^\text{mask}$ & AR$_{L}^\text{mask}$ \\
		\midrule
		COCO & 9.6 & 20.1 & 8.5 & 2.3 & 10.5 & 21.5 & 5.6 & 16.3 & 25.4 & 9.5 & 31.2 & 44.9 \\
		COCO20K & 9.8 & 20.5 & 8.4 & 2.6 & 10.5 & 21.6 & 5.6 & 16.5 & 25.6 & 9.7 & 31.6 & 44.7 \\
		LVIS & 3.7 & 7.3 & 3.2 & 1.5 & 6.9 & 12.0 & 2.1 & 7.9 & 16.1 & 6.3 & 29.2 & 41.8 \\
		\bottomrule
	\end{tabular}
	\label{tab:results_mask}
\end{table}

\begin{table}[h]
	\centering
	\tablestyle{2pt}{1.0}
	\caption{\textbf{Unsupervised object detection results on all benchmarks in this work.}}
	\begin{tabular}{l|cccccc|cccccc} 
		\toprule
		Datasets & AP$^\text{box}$ & AP$_{50}^\text{box}$ & AP$_{75}^\text{box}$ & AP$_{S}^\text{box}$ & AP$_{M}^\text{box}$ & AP$_{L}^\text{box}$ & AR$_{1}^\text{box}$ & AR$_{10}^\text{box}$ & AR$_{100}^\text{box}$ & AR$_{S}^\text{box}$ & AR$_{M}^\text{box}$ & AR$_{L}^\text{box}$ \\
		\midrule
		COCO & 12.5 & 23.8 & 11.9 & 4.0 & 13.5 & 27.7 & 6.6 & 19.8 & 32.0 & 13.2 & 38.7 & 55.8 \\
		COCO20K & 12.6 & 24.1 & 11.8 & 4.3 & 13.3 & 27.6 & 6.6 & 20.0 & 32.2 & 13.5 & 38.9 & 55.8 \\
		LVIS & 4.6 & 9.2 & 4.1 & 2.4 & 8.6 & 15.5 & 2.4 & 9.5 & 20.0 & 8.7 & 34.9 & 51.6 \\
		VOC & 20.5 & 39.1 & 19.6 & 2.7 & 8.2 & 31.5 & 15.9 & 33.2 & 44.6 & 19.3 & 36.5 & 54.5 \\
		KITTI & 8.8 & 20.8 & 6.2 & 1.2 & 6.2 & 17.4 & 6.3 & 19.7 & 29.7 & 17.4 & 26.8 & 42.1 \\
		OpenImages & 9.3 & 16.7 & 9.2 & 0.2 & 1.9 & 14.6 & 6.6 & 16.5 & 27.1 & 4.1 & 19.6 & 34.6 \\
		Objects365 & 11.2 & 22.6 & 9.8 & 2.6 & 10.5 & 19.1 & 2.8 & 15.0 & 32.0 & 11.7 & 34.6 & 45.9 \\
		\bottomrule
	\end{tabular}
	\label{tab:results_box}
\end{table}

\begin{figure}[h]
	\centering
	\includegraphics[height=1.0\textheight]{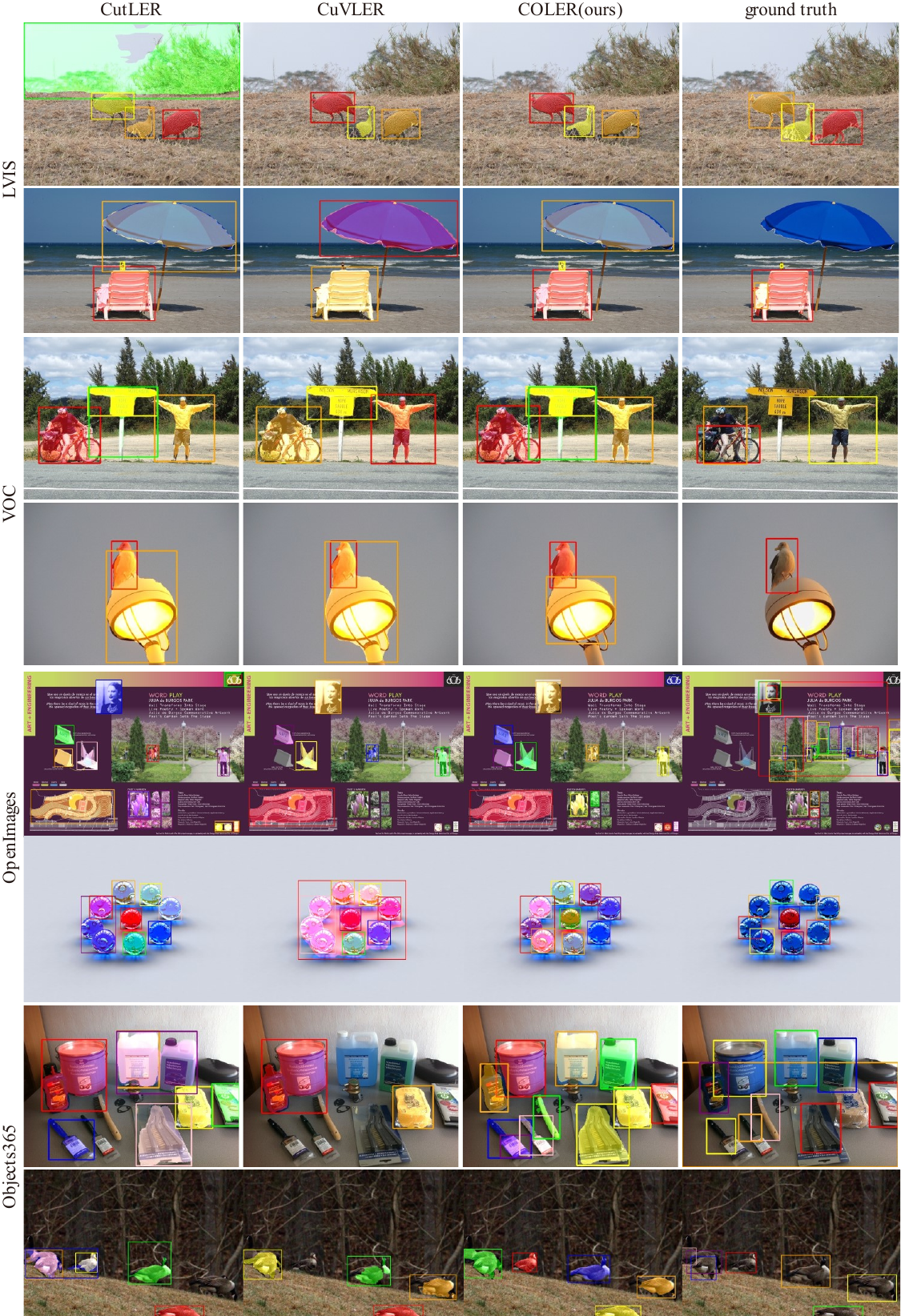}
	\caption{\textbf{Qualitative comparison of our COLER previous SOTA methods on LVIS, VOC, OpenImages, and Objects365.}}
	\label{fig:more_results}
\end{figure}

\clearpage
\begin{figure}[h]
	\centering
	\includegraphics[width=1.0\linewidth]{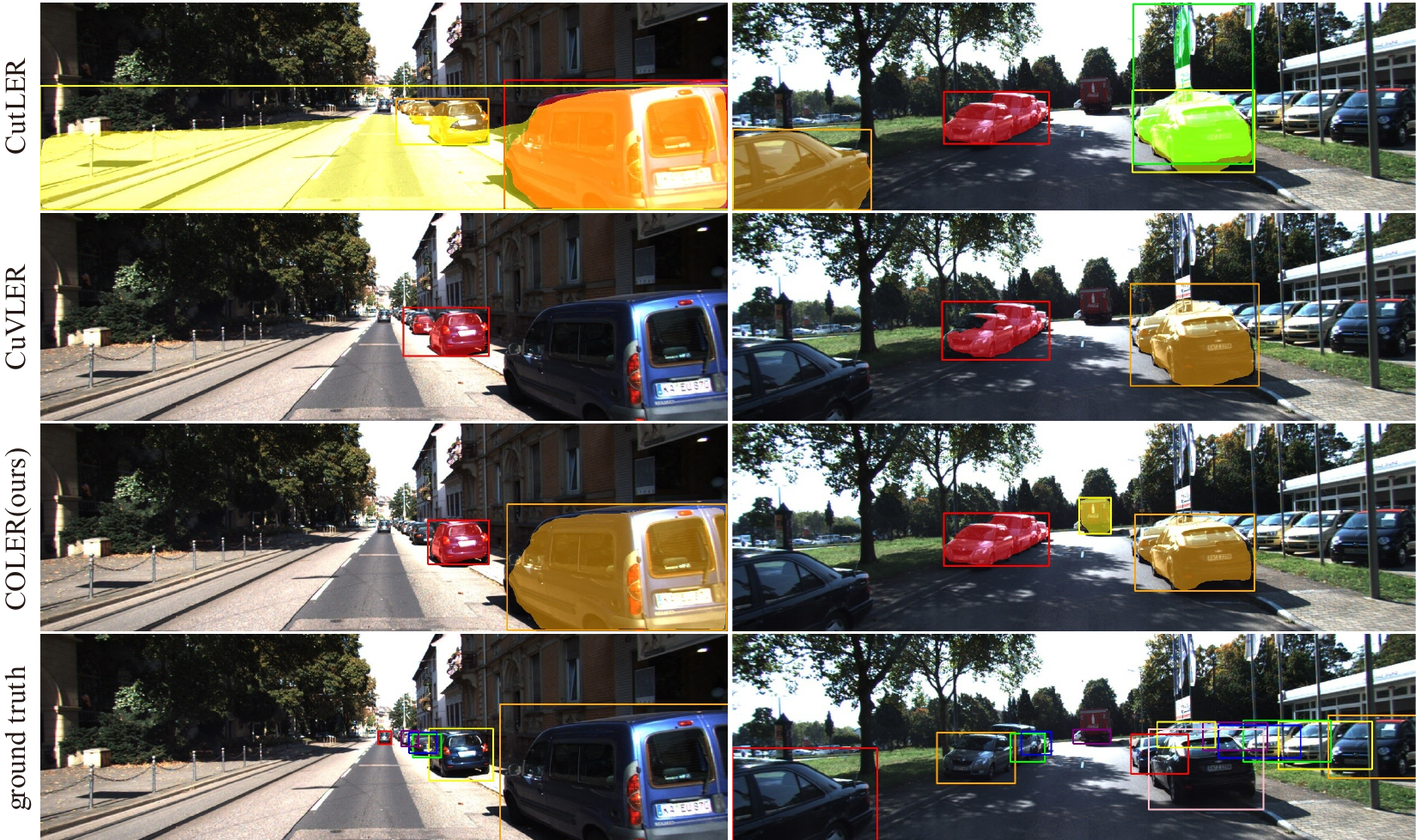}
	\caption{\textbf{Qualitative comparison of our COLER with previous SOTA methods on KITTI.}}
	\label{fig:kitti_result}
\end{figure}

\subsection{Other Visualizations}

\Figref{fig:cutonce_details} shows a visualization of the intermediate computation process of \textit{CutOnce's boundary enhancement module}. Obviously, the mechanism of this module is easy to understand and shows immediate effectiveness.

\Figref{fig:more_objects_cutonce} demonstrates the strong capability of CutOnce in segmenting multiple objects, which previous methods were unable to detect in such quantity.

\tabref{tab:results_mask} and \tabref{tab:results_box} present the zero-shot evaluation results on unsupervised instance segmentation and object detection tasks across various datasets, respectively.

\Figref{fig:kitti_result} and \Figref{fig:more_results} show additional visualization results of our COLER method compared to previous state-of-the-art approaches.
These figures only display predicted results with a confidence score of \textit{no less than 0.5} (ground truth is excluded).